\documentclass{article}
\usepackage{ijcai22}

\usepackage{times}
\usepackage{soul}
\usepackage{url}
\usepackage[hidelinks]{hyperref}
\usepackage[utf8]{inputenc}
\usepackage[small]{caption}
\usepackage{graphicx}
\usepackage{amsmath}
\usepackage{amsthm}
\usepackage{booktabs}
\usepackage{algorithm}
\usepackage{algorithmic}
\usepackage{multirow}
\title{When Performance is not Enough - A Multidisciplinary View on Clinical Decision Support}

\author{
Roland Roller$^{1,2}$
\and
Klemens Budde$^{2}$\and
Aljoscha Burchardt$^1$\and
Peter Dabrock$^{3}$\and
Sebastian Möller$^{1}$\and
Bilgin Osmanodja$^{2}$ \and
Simon Ronicke$^{2}$\and
David Samhammer$^{3}$\and
Sven Schmeier$^{1}$
\affiliations
$^1$German Research Center for Artificial Intelligence (DFKI), Germany\\
$^2$Charité - Universitätsmedizin Berlin, Germany\\
$^3$University Erlangen-Nürnberg (FAU), Germany
}

\begin{document}

\maketitle


\begin{abstract}
Scientific publications about machine learning in healthcare are often about implementing novel methods and boosting the performance - at least from a computer science perspective. However, beyond such often short-lived improvements, much more needs to be taken into consideration if we want to arrive at a sustainable progress in healthcare. What does it take to actually implement such a system, make it usable for the domain expert, and possibly bring it into practical usage? Targeted at Computer Scientists, this work presents a multidisciplinary view on machine learning in medical decision support systems and covers information technology, medical, as well as ethical aspects. Along with an implemented risk prediction system in nephrology, challenges and lessons learned in a pilot project are presented.

\end{abstract}

\section{Introduction}

The United Nations’ Sustainable Development Goal 3 “Good Health and Well-being” aims to ensure healthy lives and promote well-being at all ages, e.g., by increasing life expectancy and improving access to physicians. AI-based tools, such as intelligent decision support systems, are expected to contribute to this goal, as they promise to increase diagnostic precision and improve monitoring medical treatments. Even though the healthcare sector comes with many well justified constraints regarding the use of data or the design of experiments, it can be considered one of the drivers of applied research in AI and machine learning (ML) in particular. In the past, many scientific publications in ML evolved around the introduction of novel methods and boosting performance, e.g., of classification or prediction tasks based on readily available data \cite{Tan2021,Qiao2019,Karthikeyan2021}.

While implementing methods for the best possible system performance is important and necessary, this is only the first step and sometimes it means harvesting the low-hanging fruit. Many fundamental research questions follow directly when the goal is to make practical use of such a system in healthcare: Can it be further developed without medical expertise and if not, how and where does it come into play? How can the evolving socio-technical system be evaluated? How can we, e.g., assess criteria such as trust or transparency? Does the relationship between medical experts and patients change when the system enters the scene? It is clear that these questions cannot be answered by computer scientists alone and likewise they cannot be answered by medical or ethical experts on their own without taking technical expertise into consideration.

In order to move a step forward, this article is written for Computer Scientists working (and particularly starting to work) in the intersection of AI and medicine. Based on a risk prediction model implemented into an intelligent decision support system that we have developed in the context of nephrology, we discuss the main challenges and lessons learned from a technical perspective. Doing so, we take medical and ethical  perspectives into account,  and try to shed light on aspects that are  worth taking into consideration to get beyond a simple performance gain. The main contribution of this work are the learnings we made during the development of our risk prediction system.

\section{A Use Case in Nephrology}

The risk prediction model we present here was developed in  a project by some of the authors of this paper, namely the computer scientists and the clinical experts. The main goal of our risk prediction model was the detection of kidney transplant patients at risk of an infection within the next 90 days. Kidney transplantation requires immunosuppressive medication, and while too little immunosuppressive medication can lead to rejection and transplant failure, the majority of patients suffer from the opposite - infections. Severe bacterial infections are strongly associated with a very high CRP (C-reactive protein) level in the blood, e.g., a CRP level increases from a normal value below 5 to above 100 mg/L.   

Baseline for our work was a real and noisy patient database of a kidney transplant center, consisting of demographics, lab values, vital parameters, medications, clinical notes etc. As transplanted patients need to come to the hospital for checkups 3-4 times a year, the database contains a lot of valuable information on a fine-grained level over many years.

\begin{figure*}[!th]
\begin{center}
\includegraphics[scale=0.3]{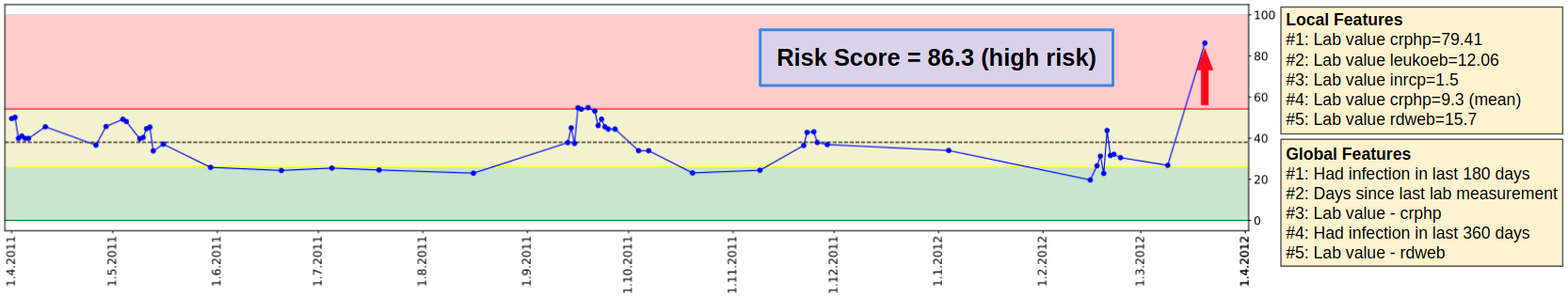} 
\caption{Simplified version of the dashboard including risk score, risk graph over time and relevant features.}
\vspace{-3mm}
\label{dashboard}
\end{center}

\end{figure*}

For our experiments data from 2009-2019 was considered, involving 1500 different patients with altogether approximately 100k data points. We define a data point as a point in the life of a patient documented in the database. Each time a new entry is made in the patient database for that patient, a new data point is created. Given a data point, including previous information until that point in time, the risk prediction model makes an estimation of how likely an infection occurs within a given time frame. About 12\% of the data points include an infection within the next 90 days.

\textbf{Infection} was defined based on CRP levels in the blood. Since CRP elevations can be caused by other non-infectious causes (e.g. rheumatic diseases or cancer) and in order to focus on relatively severe infections a threshold of 100 mg/L has been set for endpoint definition.

For our implementation we use a Gradient Boosted Regression Tree (GBRT), which can be trained quickly, also on less performant infrastructure. Overall, our model was intended to be a baseline for a first prototype. It was developed within the hospital, given the available infrastructure, in close collaboration with medical experts. Before training, the data was preprocessed, transformed and filtered. The final model uses about 300 different features, consisting of lab values, vital parameters, diagnoses, medications, demographics etc. The resulting model, as well as further descriptions about the data and usage of the model, can be made available on request.
\begin{table}[ht!]
\centering
\begin{tabular}{lll}
\toprule
\textbf{Days} & \textbf{ROC} & \textbf{PRC}  \\ 
\midrule

90 & \textbf{0.80} & 0.41 \\ 
180 &0.79 & 0.46 \\ 
360 & 0.77 & \textbf{0.50} \\ 
\bottomrule
\end{tabular}
\caption{Result on 10 fold cross-validation to predict the endpoint “infection”, according to Area under Receiver Operator Curve (ROC) and Area under Precision Recall Curve (PRC).}
\vspace{-4mm}
\label{tab_internal_validation}
\end{table}

\subsection{Internal Validation}
Within an internal validation we evaluated our system using a 10-fold cross validation on the historic (retrospective) patient data. Each time, the data was randomly split into training, development and test using a split of 70/15/15. The split was conducted on a patient level, to ensure that no data point of the same patient will occur across different splits. In addition to the endpoint prediction for the next 90 days, we also explore the prediction within the next 180 and 360 days. Even though Area under Receiver Operator Curve (ROC) is frequently used for the evaluation of this kind of tasks \cite{Topol2019}, it is known that ROC cannot deal very well with unbalanced data \cite{Branco2016}, which is often the case in a clinical setup. Therefore for our evaluation we use ROC, as well as Area under Precision Recall Curve (PRC).

The results of the internal validation are presented in Table 1. For the intended prediction within 90 days, the ROC score is 0.80 and the PRC score 0.41. In comparison with the longer time frames, the ROC score is a bit higher for the event in the near future while the PRC score increases for the long term predictions.

\subsection{Experiment: Machine Learning vs Physicians}
Based on the internal validation, it was difficult to assess the quality and efficiency of the system as to the best of our knowledge no related work addressing this problem existed. Therefore it was also not clear if this model would be useful for medical experts in some way or other. Thus, instead of focusing on the further optimization of our machine learning model, we decided to find out how well physicians can solve the given task, and if our approach might have a benefit for them already.

In the course of this, we conducted an experiment, consisting of two parts: In the first part, physicians received a data point of a patient, together with all information up to this time, and had to make a risk estimation (0-100\%) if an infection as defined above will occur within the next 90 days. In the second part the physician additionally received information of the risk prediction model, in the form of a dashboard (see Figure \ref{dashboard}). The dashboard depicts the risk score over time within a graph, mapped to a traffic light system, indicating if the patient has got a low, medium or high risk of an infection, along with local and global features of the risk prediction model. Whereas global features are generally important features for the overall model, local features are those which had the strongest impact on the current prediction of the system. The cutoff values for the traffic light system were calculated by using the predictions on a separate development set. While the dashed line in the middle represents the threshold for the optimal F1 score, the yellow area is defined by F2, which highlights recall over precision, and the red area is defined by F0.5, which highlights precision over recall.  

While automatic risk estimations can be generated within seconds, humans require time to carry out this task. This fact indeed limited our possibilities for human evaluation. Overall, our study included 120 data points of 120 different patients (38 infections, 82 without infections). 8 physicians participated in our study, four experienced (senior) and four assistant (junior) physicians. Within each part of the study, each physician received 15 data points to analyze. The study was accepted by the ethical advisory board of the hospital, and according to the staff council each physician had up to 30 minutes time to examine the data of each patient. Each physician received some basic introduction about the dashboard, information about how the machine learning model works, as well as its performance on the internal validation study. Finally, the result of both parts of the study are then compared to the performance of the risk prediction system itself. In the following we refer to the first part of the study as \textbf{MD} (medical doctor) and the second part as \textbf{MD+AI} (medical doctor including machine learning support). The machine learning component itself will be referred to as \textbf{AI}. 

For the experiment we re-trained the model, and made sure that no patient of the test set occurs within the training and development set. The evaluation of the experiment is carried out according to ROC and PRC, as well as sensitivity (recall), specificity (true negative rate), and positive predictive value (precision) using different cutoff-thresholds.

\begin{table}[ht!]
\small
\centering
\begin{tabular}{lllllll}
\toprule
 & \textbf{ROC} & \textbf{PRC} & \textbf{SEN} & \textbf{SPEC} & \textbf{PPV} & \textbf{thrs} \\ 
\midrule

\multirow{3}{*}{AI} & \multirow{3}{*}{0.719} & \multirow{3}{*}{0.567} & 0.754 & 0.718 & 0.264 & F2 \\ 
 &  &  & 0.580 & 0.844 & 0.333  & F1 \\ 
 &  &  & 0.329 & 0.940 & 0.424 & F0.5 \\ 
 \midrule
MD & 0.630 & 0.474 & 0.448 & 0.689 & 0.406 & $>$50\% \\
\midrule
MD+AI & 0.622 & 0.469 & 0.368 & 0.829 & 0.500 & $>$50\% \\
\bottomrule
\end{tabular}
\caption{Result of the physicians predicting infections within the next 90 days - without (MD) and with (MD+AI) decision support, in comparison to the automatic risk prediction (AI). Evaluation is carried out using ROC (Area under Receiver Operator Curve) and PRC (Area under Precision Recall Curve), as well as SEN (sensitivity; recall), SPEC (specificity; true negative rate), and PPV (positive predictive value; precision) defined by different thresholds (thrs).}
\label{tab_main_experiments}
\end{table}

The results presented in Table 2 show that overall the task is not too easy for physicians, considering the ROC score of 0.63. Moreover, the results indicate that physicians receiving the additional automatic decision support do not increase performance. On the other hand, the risk prediction system outperforms the physicians in both parts of the study regarding ROC and PRC. 

While the ROC and PRC results of AI show a promising improvement in comparison to the physicians, this does not tell us if patients would be labeled as critical (infection will occur) or uncritical. Thus we use different cutoff values to map from the probability estimation of the physicians and the regression score of the risk prediction system, to either 1 (infection) and 0 (no infection). In this way, we can evaluate the risk prediction system, as well as the physicians according to sensitivity, specificity, and positive predictive value. The results in Table 2 show that AI achieves the highest sensitivity, therefore finds the largest number of critical patients, depending on the selection of the cutoff value. Conversely, all results present only a moderate performance according to the positive predictive value (PPV). In the case of MD+AI for instance every second precision would be a true positive infection prediction, while the sensitivity is only about 37\%. Similarly it behaves for the others, which achieve a higher sensitivity, but it comes at the price of a reduced PPV.

Table 3 compares the ROC performance of the two subject groups in both parts of the study. The table shows that senior MDs are better than junior MDs in solving the task. Moreover, the table indicates that junior MDs increase in performance, together with the automatic decision support, while senior MDs decrease.

\begin{table}[ht!]
\centering
\begin{tabular}{lll}
\toprule
\textbf{} & \textbf{Junior MD} & \textbf{Senior MD}  \\ 
\midrule

MD & 0.5781 & \textbf{0.6772} \\ 
MD+AI & \textbf{0.6398} & 0.6149 \\ 
\bottomrule
\end{tabular}
\caption{ROC performance of the two different subject groups, without (MD) and with (MD+AI) automatic decision support.}
\label{tab_experiment_subject_group}
\end{table}
\vspace{-4mm} 

\section{Learnings made during the Development}

\paragraph{Development constraints}
In general, many circumstances affected the overall development of the risk prediction model. The development had to be conducted within the hospital infrastructure and network, due to the sensitivity of the data. This meant that limited computing power was available, partially admin rights of the working computer were missing and access to the internet was restricted. 

\paragraph{Data constraints}
Real clinical data tends to be noisy and to have grown over time. In our case, some older data did not have the same quality as more recent data points. Certain database fields were not used anymore or were used only occasionally. This means, much domain and expert knowledge was required to interpret and subsequently filter and normalize the data, including lab values which occurred in different units and eventually even were assessed with different lab methods over time. 

\paragraph{Never trust a single score}
A somewhat trivial advice has proven helpful: Never trust a single score. While a ROC score of 0.80 in the internal validation appears to be promising, it is surprising that the 180 and 360 day prediction achieve similar results, as a prediction in those long term time frames seems to be quite difficult. One reason could be that the model might be able to identify factors, which generally increase the risk for infections, rather than those predicting a particular problem for the near future. In addition, the prediction of an infection within the next 360 days, also includes events which occur already in the near future. Moreover, the PRC score for 90 days is only mediocre, showing an even stronger counterintuitive increase for longer time frames.


\paragraph{The issue with false positives}
In our scenario, depending on how the threshold is chosen, the results of sensitivity, specificity and PPV change. From a patient perspective the presented results are far away from being optimal. None of the approaches will predict all possible patients at risk, and the more critical patients can be detected the lower the PPV. Can such a setup be possibly useful for clinical practice? Particularly if the number of false positives is too large the automatic prediction might not be seen as trustworthy anymore, or might lead to “alert fatigue” \cite{Ash2007}. Reporting non-perfect machine learning scores on medical prediction tasks appears to be common, as no system provides a perfect score \cite{Tomavsev2019,Guo2019}. However, reporting results without taking the medical expertise into account, and without knowing the benefit might be not optimal.

\paragraph{Data analysis}
The usage of multiple scores is certainly beneficial, to get a good impression of the quality of a system. On the other hand looking into the data, and analyzing the estimations of single patients is essential. Is there something the automatic risk prediction can do that the physician can't do? Or can only the obvious cases be predicted? In our experiment for instance, a detailed analysis revealed, that the risk prediction system can partially detect different patients at risk, which is a strong argument for using the system as a double check for physicians.

\paragraph{Communication of results}
Even though our model outperformed physicians on the risk prediction task, the experiment showed that physicians cannot make better estimations together with automatic predictions. This shows that performance alone is not sufficient to introduce novel technologies into clinical practice. Moreover, as we have seen in Table 3, junior MDs increased in ROC performance, while senior MDs did not. This indicates that the human-machine-interaction and in particular the way how information is communicated might be essential and need to be further studied for the success of the system.

\section{Discussion and more general Learnings}

\paragraph{Communicate with relevant stakeholders}
In order to develop machine learning models on clinical data, communication is key. Frequent interactions with relevant stakeholders certainly help to understand the task, and to identify relevant factors to solve it, or to understand which information is reliable (or has to be ignored) in the database. As simple as this sounds, it is not, as for instance physicians have got a different background and so do you. It is necessary to find a “common language”. In an ongoing project not reported here, we additionally take the patient’s perspective into account. While this is undoubtedly important, it comes with another bunch of challenges.

\paragraph{Start with a simple system}
From a machine learning perspective, the selection of the underlying model has a huge influence on the outcomes. Therefore neural models seem to be the first choice. However, in most cases we recommend starting with a simple system in order to develop a solid and reliable baseline first. A model which can be quickly developed, and quickly trained, also on less powerful computing architecture - as we did. Moreover, starting generally simple, with fewer features and slowly increasing the complexity can be beneficial, as (sequential) medical data can be complex. 

\paragraph{Make use of expert knowledge}
Machine learning engineers often think that a model solely needs to be trained on large data in order to solve the problem by itself. In fact, data driven machine learning models might be able to learn new or at least different correlations and patterns as compared to physicians who can reason in terms of causation. In our setup for instance, the number of lab values appears to be a good indication that something is going wrong - conversely physicians would probably rather argue that some values are “borderline” or abnormal, therefore additional lab examinations might be needed, which has been also observed in other related work \cite{Rethmeier2020}. Although different, both observations indirectly refer to the same - more lab examinations are needed, as something does not seem to be okay. In any way, we recommend taking the opinion of physicians into account (ideally from the beginning) and examine if their suggestions about relevant information/parameters contribute to the system performance. For instance, is the suggested information covered by the data going into the model, or can the information actually properly be represented by the model (e.g. fluctuations of particular lab values)?

\paragraph{Use Case/Target Definition}
For every research project in medicine, endpoint definition is crucial. Endpoints need to be precise and clinically relevant. When using clinical routine data to train machine learning algorithms, endpoints need to be defined from preexisting data, which can be tricky. While endpoints in prospective clinical trials are explicitly assessed by performing diagnostic tests at a certain time point defined in a study protocol (e.g. kidney biopsy one year after study inclusion), this is not the case for retrospective trials using clinical routine data. Hence, it is not advisable to use endpoints, for which diagnostic tests are not regularly performed. In our example, to predict infections, we chose a laboratory-based definition, because for the majority of data points in our dataset, this laboratory value is available and always will be determined in case of severe infection. On the contrary, it would be much more problematic to predict a rare event, which requires a specific diagnostic examination that is not routinely performed. Every prediction based on such data will be imprecise, because the diagnosis is often missed or at least delayed by the physicians themselves. It is important to keep in mind that no AI tool can detect cases it has not learned to detect. Even with that in mind, defining a broad endpoint such as infection using a single laboratory parameter can be too simplistic and again not optimal for the clinicians. Specifically, our definition ignores less severe, and non-bacterial infections, but includes certain rare, non-infectious reasons for increased CRP-value. Therefore, to prevent the necessity of redefining endpoints later in the model development, it is again advisable to consult several clinicians about their opinion about specific endpoint definition, as the endpoint definition not only has to meet clinical usability but also has to be accepted by the physician. 

\paragraph{Lack of comparability}
The evaluation of a method is essential. On the other hand, comparability and reproducibility is a major problem for AI in healthcare, as datasets are often not available. Each data set has its own characteristics. Achieving good results on one dataset is not a guarantee that you also achieve similar results on a different one, addressing the same problem. One reason for this, if the underlying dataset (cohort) is selected slightly differently.

\paragraph{Setting of the study and choice of study subjects}
In addition to performance evaluation, most medical problems involve at least a medical expert - and potentially an AI-based system - for resolving the task. Thus, the interplay between medical experts and AI-based systems needs to be considered in the evaluation. Human-machine interaction is just one aspect of this interplay. Importantly, the set-up of the experiments in which humans and the system collaborate, the information that is given to the human participants about the AI-based system, the setting of the study, etc. are all factors which are likely to impact the results. Thus, a simple comparison between human and machine baseline is not really helpful for estimating how such a sociotechnical system would perform in practice. If the patient’s view is to be taken into account, depending on the state of the patients, it can be advisable to also include nursing staff, relatives or even third parties such as patient representative organizations.

\paragraph{Model explanations}
The “explanations” of our system are based on local and global features –  a simple and pragmatic solution. In many cases this is a particular lab value, the presence of a diagnosis, intake of a medication or the time since the last transplantation. In contrast, in previous work we explored how physicians would actually justify their risk estimations. Those justifications were short paragraphs consisting of a few sentences and partially contain similar information as the relevant features of the ML system. However, justifications contain, aside from pure facts, also descriptions about fluctuation and tendencies of certain parameter (“lab value is raising since the last few months”), negations/speculations (e.g. explicit absence of symptoms) , additional world knowledge (“lab value above norm”, “patient does not seem to be adherent”) and interpretations (“patient is fit”). Overall, medical decisions do not only depend on medical knowledge/experience and parameters provided in a database, but also from conversions to the patient or nursing staff. This information is only partially contained in clinical databases, if then only in clinical text. Moreover, we experienced that human justifications strongly vary, not only depending on task and risk score, but also depending on the individual physician. This indicates that a “one-size-fits-all” perfect system explanation will probably not exist, as each physician might have their own   preferences. As understanding the system recommendation is essential to build up trust, the topic of communicating and justifying system decisions is  still an open research domain - although much work is currently conducted in the context of XAI (explainable AI) \cite{Markus2021}.

\paragraph{A customizable user interface}
Another relevant aspect we experience in our experiment is the wish of a personalization of such an automatic decision support system, which might be grounded in the need for professional communication mentioned above. The current approach is static, it is optimized according to ROC, and presents results within a Dashboard. However, at the end of our experiments, physicians expressed the interest to interact and modify the system according to their preferences. For instance by excluding parameters, or increasing the relevance of some aspects. We assume that an automatic and interactive decision support system might help to easier understand the technology behind, and understand the potential, as well as limitations of such a system. In this way, a more efficient supportive system could be created, according to the wishes and needs of the treating physician, e.g. targeting only on high precision predictions, or achieving a larger recall. This might help to build up more trust in this technology. Techniques that are discussed under the heading of XAI like counterfactual explanations or the display of similar cases from the training data may further improve the usefulness of the system here.

\paragraph{From Model to Usage}
From a clinical perspective, many applications of AI represent improvements on a technological level, but approach problems that are either mastered by human physicians or can and have been solved with classical statistical methods as well. For AI researchers working with medical data, the long-term goal should be to arrive at a system that is able to improve patient outcomes or reduce physicians’ workload. Only then will it be considered beneficial by evidence-based medicine or by physicians in daily practice. Great performance alone will not be enough for that. We think that systems aiming above all for successful translation into clinical practice will have a higher impact than very performant prediction tools for a suboptimal use case. Unfortunately, this is usually more difficult to achieve with respect to data collection, preparation and outcome definition. Additionally, structural barriers of research funding limit the possibility to follow long-term projects.

Nevertheless, for a tool to be eventually implemented in a clinical setting, design questions should be addressed carefully in the beginning. Specifically, use cases should be defined before the development of a tool by performing user research involving physicians, nurses, or patients at the beginning. They should be selected based on existing weaknesses in clinical settings and consider the problems and limitations of clinical work and reasoning. By choosing such tasks, the potential benefit can be maximized and is easier to demonstrate in controlled trials studying the system’s impact. Finally, a close interaction with the end user during implementation seems crucial as the end users have to be trained not only on practical aspects but also on the strengths and weaknesses of the model in order to make an informed treatment decision.

\subsection{Connecting different Stakeholders}
It has become standard practice for better AI decision support systems to claim that they are aligned with the concept of trustworthy AI. Numerous criteria have been developed for this purpose, for which various reviews are now available \cite{Morley2021,Jobin2019}. If one also takes the frame of the HLEG as a reference, one can say: It is important that any AI-System should be compatible with applicable laws, meets ethical standards, and does not entail unforeseen side effects \cite{HLEG}. However, trust is not a concept that is valid once achieved but must be constantly maintained \cite{Braun2021b}. Therefore, it is crucial how the path from principles to practice is designed. Formally, approaches of co-creation AI \cite{Zicari2021} are well suited for this challenge. But also terms of content need to be unfolded. For this purpose, we decided in our experiment to analyze and evaluate attitudes of the physicians involved.

Of particular interest for AI-driven support systems in the clinic seems to be the question of how interaction processes change through the introduction of such and how normative concepts like trust, responsibility, and transparency have to be rethought in a new and interrelated way \cite{Braun2021}. To approach these questions in the context of our experiments, qualitative interviews were conducted with the physicians at the end of the study. We decided to use semi-structured expert interviews as the data collection method. The interviews were intended both to obtain an evaluation of the conducted case study and to find out what impact the introduction of AI-driven support systems have on the interviewed physicians. The evaluation is mainly relevant for the further development of the tested system while other estimations of the physicians can help to draw conclusions for further ethical discussions.

The results are reassuring. The physicians tell us how they used the system, to what extent they trusted the assessment and what suggestions they have for improvement. Especially for the design of such experiments, these statements are of major importance. For example, it became clear that some of the physicians followed a certain procedure they gave themselves, e.g., by first noting their own prediction and then checking the system result (“second opinion”) or by first looking at the system result and then challenging it with the available evidence. When designing further experiments, we can now better decide if we want to impose a certain procedure or not.

At the same time, the physicians report on the complexity of clinical decision-making in general. They confirm trust to be a prerequisite for being able to make decisions. For an AI-driven support system to be transferred to clinical practice at all, it must therefore be trusted. According to the physicians, the evaluations of the systems must be explainable for this. Not completely, but in a way that the physicians are able to present the application of the system to the patient, to whom responsibility is always borne.

Without being able to go into more detail about the content of the interviews at this point, we want to emphasize the positive experience that took place with the qualitative survey accompanying the case study. What becomes clear is that through the interviews a connection is made between the profession of the physicians and the researchers. Following the thesis of \citeauthor{Noordegraaf2020} that professions have to be more and more in touch with their environment, especially due to technical challenges, we can say from our perspective that the qualitative data collection is definitely able to be a link between the physicians and their environment. Our recommendation to future research projects is therefore that they should be accompanied by such or similar studies. These should not only serve as an evaluation of the individual projects, but also as part of designing AI decision support systems and as a contribution of the communication process which must be part of the transformations that already take place.

\vspace{-3mm}
\section{Related Work}
Due to limited space, here, we just briefly draw comparison to the most relevant work. The work of \citeauthor{Sutton2020} provides a good introduction into the world of clinical decision support and discusses some of our learnings of our work, such as the problem of too many false positives. \citeauthor{Zicari2021} discuss the development of an ethically aligned co-design to ensure trustworthiness. However, the work remains more theoretical, with less and concrete practical advice as the work reports on an early developmental stage and thus misses the aspect of a real system and experiments to discuss. In \citeauthor{Amann2020}, the authors came to a similar conclusion as we do, that performance alone is not enough, and therefore stress the importance of explanabilty. The authors discuss this topic from multiple perspectives, but provide only an overview about the topic without getting too much into detail. \citeauthor{Bruckert2020}, again a high level and interdisciplinary work, present a roadmap to develop comprehensive, transparent and trustful systems, considering the “black-box-nature” of many machine learning approaches. However, the work targets mainly on the aspect of explainability and interaction, and does not validate the ideas within a particular experiment.  

Although not too recent, the work of \citeauthor{Topol2019} provides an overview about different medical fields and the most relevant work regarding developments of AI algorithms. The authors also present publications from different medical fields in which the performance of AI methods have been compared to  the performance of doctors - nephrology is not listed here. Moreover, the author highlights that AI can enhance the patient-doctor relationship - a perspective which has not been analyzed in detail in our work reported here. Finally, the work of \citeauthor{Ho2021} arrives at a similar conclusion to ours, that the inclusion of multiple stakeholders is necessary to develop AI technology in healthcare. 

Concluding, there is a large range of publications addressing the aspect of automatic AI-driven clinical decision support (or particular aspects of it), its challenges and coming to partially similar conclusions as we do. However, this work discusses multiple aspects developing such a system from a rather technical aspect, meant for Computer Scientists, but taking also other perspectives into account which are necessary. Moreover, the discussion does not remain theoretical, as we substantiate it with challenges and learnings we have experienced along a use case in nephrology.

\section{Conclusion}

This work presented an automatic risk prediction system to predict severe infections after kidney transplantation within the next 90 days. Within a small study we compared the performance of our system to the performance of physicians with and without the support of our system. Main findings are that our risk prediction system outperforms physicians on our small dataset, physicians do not improve together with the automatic decision support, only junior physicians do. Although from a technical point of view, our system achieves promising results, our experiments leave us with more open questions than before, such as what does a successful decision support system need to look like to achieve the goal to actually help physicians treating their patients in a better way? Along with our experiments we discussed the challenges and lessons learned from a multidisciplinary view. Targeted to Computer Scientists new in the field of AI in healthcare, the main contribution of this work is to highlight that pure performance is not enough to set up a practical tool to support clinicians. Moreover, the presented challenges indicate particularly the need in multidisciplinary work with different stakeholders, as well as shaping the AI-human interaction on different levels.

\section*{Acknowledgements}%
This research was supported by the German Federal Ministry of Education and Research (BMBF) through the project vALID (01GP1903A).


\bibliographystyle{named}
\bibliography{ijcai22}

\begin{thebibliography}{}

\bibitem[\protect\citeauthoryear{Amann \bgroup \em et al.\egroup
  }{2020}]{Amann2020}
Julia Amann, Alessandro Blasimme, Effy Vayena, Dietmar Frey, and Vince~I Madai.
\newblock Explainability for artificial intelligence in healthcare: a
  multidisciplinary perspective.
\newblock {\em BMC Medical Informatics and Decision Making}, 20(1):1--9, 2020.

\bibitem[\protect\citeauthoryear{Ash \bgroup \em et al.\egroup
  }{2007}]{Ash2007}
Joan~S Ash, Dean~F Sittig, Emily~M Campbell, Kenneth~P Guappone, and Richard~H
  Dykstra.
\newblock Some unintended consequences of clinical decision support systems.
\newblock In {\em Amia annual Symposium proceedings}, volume 2007, page~26.
  American Medical Informatics Association, 2007.

\bibitem[\protect\citeauthoryear{Branco \bgroup \em et al.\egroup
  }{2016}]{Branco2016}
Paula Branco, Lu{\'\i}s Torgo, and Rita~P Ribeiro.
\newblock A survey of predictive modeling on imbalanced domains.
\newblock {\em ACM Computing Surveys (CSUR)}, 49(2):1--50, 2016.

\bibitem[\protect\citeauthoryear{Braun \bgroup \em et al.\egroup
  }{2021a}]{Braun2021b}
Matthias Braun, Hannah Bleher, and Patrik Hummel.
\newblock A leap of faith: is there a formula for “trustworthy” ai?
\newblock {\em Hastings Center Report}, 51(3):17--22, 2021.

\bibitem[\protect\citeauthoryear{Braun \bgroup \em et al.\egroup
  }{2021b}]{Braun2021}
Matthias Braun, Patrik Hummel, Susanne Beck, and Peter Dabrock.
\newblock Primer on an ethics of ai-based decision support systems in the
  clinic.
\newblock {\em Journal of medical ethics}, 47(12):e3--e3, 2021.

\bibitem[\protect\citeauthoryear{Bruckert \bgroup \em et al.\egroup
  }{2020}]{Bruckert2020}
Sebastian Bruckert, Bettina Finzel, and Ute Schmid.
\newblock The next generation of medical decision support: A roadmap toward
  transparent expert companions.
\newblock {\em Frontiers in artificial intelligence}, page~75, 2020.

\bibitem[\protect\citeauthoryear{Guo \bgroup \em et al.\egroup
  }{2019}]{Guo2019}
Xu~Guo, Han Yu, Chunyan Miao, and Yiqiang Chen.
\newblock Agent-based decision support for pain management in primary care
  settings.
\newblock In {\em IJCAI}, pages 6521--6523, 2019.

\bibitem[\protect\citeauthoryear{HLE}{2022}]{HLEG}
{High-Level Expert Group on Artificial Intelligence: Ethics guidelines for
  trustworthy AI. European Commission.}
\newblock
  \url{https://ec.europa.eu/digital-single-market/en/news/ethics-guidelines-trustworthy-ai},
  2022.
\newblock Accessed: 2022-02-11.

\bibitem[\protect\citeauthoryear{Ho and Caals}{2021}]{Ho2021}
Calvin Wai-Loon Ho and Karel Caals.
\newblock A call for an ethics and governance action plan to harness the power
  of artificial intelligence and digitalization in nephrology.
\newblock In {\em Seminars in nephrology}, volume~41, pages 282--293. Elsevier,
  2021.

\bibitem[\protect\citeauthoryear{Jobin \bgroup \em et al.\egroup
  }{2019}]{Jobin2019}
Anna Jobin, Marcello Ienca, and Effy Vayena.
\newblock The global landscape of ai ethics guidelines.
\newblock {\em Nature Machine Intelligence}, 1(9):389--399, 2019.

\bibitem[\protect\citeauthoryear{Karthikeyan \bgroup \em et al.\egroup
  }{2021}]{Karthikeyan2021}
Akshaya Karthikeyan, Akshit Garg, PK~Vinod, and U~Deva Priyakumar.
\newblock Machine learning based clinical decision support system for early
  covid-19 mortality prediction.
\newblock {\em Frontiers in public health}, 9, 2021.

\bibitem[\protect\citeauthoryear{Markus \bgroup \em et al.\egroup
  }{2021}]{Markus2021}
Aniek~F Markus, Jan~A Kors, and Peter~R Rijnbeek.
\newblock The role of explainability in creating trustworthy artificial
  intelligence for health care: a comprehensive survey of the terminology,
  design choices, and evaluation strategies.
\newblock {\em Journal of Biomedical Informatics}, 113:103655, 2021.

\bibitem[\protect\citeauthoryear{Morley \bgroup \em et al.\egroup
  }{2021}]{Morley2021}
Jessica Morley, Luciano Floridi, Libby Kinsey, and Anat Elhalal.
\newblock From what to how: an initial review of publicly available ai ethics
  tools, methods and research to translate principles into practices.
\newblock In {\em Ethics, Governance, and Policies in Artificial Intelligence},
  pages 153--183. Springer, 2021.

\bibitem[\protect\citeauthoryear{Noordegraaf}{2020}]{Noordegraaf2020}
Mirko Noordegraaf.
\newblock Protective or connective professionalism? how connected professionals
  can (still) act as autonomous and authoritative experts.
\newblock {\em Journal of Professions and Organization}, 7(2):205--223, 2020.

\bibitem[\protect\citeauthoryear{Qiao \bgroup \em et al.\egroup
  }{2019}]{Qiao2019}
Zhi Qiao, Xian Wu, Shen Ge, and Wei Fan.
\newblock Mnn: multimodal attentional neural networks for diagnosis prediction.
\newblock In {\em Proceedings of the Twenty-Eighth International Joint
  Conference on Artificial Intelligence (IJCAI-19)}, 2019.

\bibitem[\protect\citeauthoryear{Rethmeier \bgroup \em et al.\egroup
  }{2020}]{Rethmeier2020}
Nils Rethmeier, Necip~Oguz Serbetci, Sebastian M{\"o}ller, and Roland Roller.
\newblock Efficare: Better prognostic models via resource-efficient health
  embeddings.
\newblock In {\em AMIA Annual Symposium Proceedings}, volume 2020, page 1060.
  American Medical Informatics Association, 2020.

\bibitem[\protect\citeauthoryear{Sutton \bgroup \em et al.\egroup
  }{2020}]{Sutton2020}
Reed~T Sutton, David Pincock, Daniel~C Baumgart, Daniel~C Sadowski, Richard~N
  Fedorak, and Karen~I Kroeker.
\newblock An overview of clinical decision support systems: benefits, risks,
  and strategies for success.
\newblock {\em NPJ digital medicine}, 3(1):1--10, 2020.

\bibitem[\protect\citeauthoryear{Tan \bgroup \em et al.\egroup
  }{2021}]{Tan2021}
Qingxiong Tan, Mang Ye, Grace Lai-Hung Wong, and PongChi Yuen.
\newblock Cooperative joint attentive network for patient outcome prediction on
  irregular multi-rate multivariate health data.
\newblock In {\em Proceedings of the Thirtieth International Joint Conference
  on Artificial Intelligence (IJCAI-21)}, 2021.

\bibitem[\protect\citeauthoryear{Toma{\v{s}}ev \bgroup \em et al.\egroup
  }{2019}]{Tomavsev2019}
Nenad Toma{\v{s}}ev, Xavier Glorot, Jack~W Rae, Michal Zielinski, Harry Askham,
  Andre Saraiva, Anne Mottram, Clemens Meyer, Suman Ravuri, Ivan Protsyuk,
  et~al.
\newblock A clinically applicable approach to continuous prediction of future
  acute kidney injury.
\newblock {\em Nature}, 572(7767):116--119, 2019.

\bibitem[\protect\citeauthoryear{Topol}{2019}]{Topol2019}
Eric~J Topol.
\newblock High-performance medicine: the convergence of human and artificial
  intelligence.
\newblock {\em Nature medicine}, 25(1):44--56, 2019.

\bibitem[\protect\citeauthoryear{Zicari \bgroup \em et al.\egroup
  }{2021}]{Zicari2021}
Roberto~V Zicari, Sheraz Ahmed, Julia Amann, Stephan~Alexander Braun, John
  Brodersen, Fr{\'e}d{\'e}rick Bruneault, James Brusseau, Erik Campano, Megan
  Coffee, Andreas Dengel, et~al.
\newblock Co-design of a trustworthy ai system in healthcare: deep learning
  based skin lesion classifier.
\newblock {\em Frontiers in Human Dynamics}, page~40, 2021.

\end{thebibliography}

\end{document}